\documentclass[letterpaper]{article} 
\usepackage{aaai24}  
\usepackage{times}  
\usepackage{helvet}  
\usepackage{courier}  
\usepackage[hyphens]{url}  
\usepackage{graphicx} 
\usepackage{natbib}  
\usepackage{caption} 
\usepackage{algorithm}
\usepackage{algorithmic}
\usepackage{amsfonts}
\usepackage{bm}
\usepackage{booktabs}
\usepackage{soul}
\usepackage{amssymb}
\usepackage{subcaption}
\usepackage{multirow}
\usepackage{makecell}
\usepackage{xcolor}

\usepackage{newfloat}
\usepackage{listings}
\DeclareCaptionStyle{ruled}{labelfont=normalfont,labelsep=colon,strut=off} 
\floatstyle{ruled}
\newfloat{listing}{tb}{lst}{}
\floatname{listing}{Listing}
\title{AutoMixer for Improved Multivariate Time-Series Forecasting on Business and IT Observability Data}
\author{
    Santosh Palaskar\textsuperscript{\rm 1}\thanks{The author was in IBM Research while the work was done.},
    Vijay Ekambaram\equalcontrib\textsuperscript{\rm 2},
    Arindam Jati\equalcontrib\textsuperscript{\rm 2},
    Neelamadhav Gantayat\textsuperscript{\rm 2},
    Avirup Saha\textsuperscript{\rm 2},
    Seema Nagar\textsuperscript{\rm 2},
    Nam H. Nguyen\textsuperscript{\rm 2},
    Pankaj Dayama\textsuperscript{\rm 2},
    Renuka Sindhgatta\textsuperscript{\rm 2},
    Prateeti Mohapatra\textsuperscript{\rm 2},
    Harshit Kumar\textsuperscript{\rm 2},
    Jayant Kalagnanam\textsuperscript{\rm 2},
    Nandyala Hemachandra\textsuperscript{\rm 1},
    Narayan Rangaraj\textsuperscript{\rm 1}
}
\affiliations{
    \textsuperscript{\rm 1}Indian Institute of Technology Bombay\\
    \textsuperscript{\rm 2}IBM Research\\
   
}

\begin{document}

\newcommand{\ie}{i.e., }
\newcommand{\eg}{e.g., }
\newcommand{\etc}{\textit{etc.}}
\newcommand{\viz}{\textit{viz. }}
\newcommand{\etal}{\textit{et. al.}}
\newcommand{\x}{\mathbf{x}}
\newcommand{\z}{\mathbf{z}}
\newcommand{\zprime}{\mathbf{z^\prime}}
\newcommand{\yprime}{\mathbf{y^\prime}}
\newcommand{\xtilde}{\mathbf{\tilde{x}}}
\newcommand{\y}{\mathbf{y}}
\newcommand{\yhat}{\mathbf{\hat{y}}}

\newcommand{\biz}{BizITObs}
\newcommand{\noteAS}[1]{\textcolor{blue}{as: #1}}

\maketitle

\begin{abstract}

The efficiency of business processes relies on business key performance indicators (Biz-KPIs), that can be negatively impacted by IT failures. Business and IT Observability (BizITObs) data fuses both Biz-KPIs and IT event channels together as multivariate time series data. Forecasting Biz-KPIs in advance can enhance efficiency and revenue through proactive corrective measures.
However, BizITObs data generally exhibit both useful and noisy inter-channel interactions between Biz-KPIs and IT events that need to be effectively decoupled. This leads to suboptimal forecasting performance when existing multivariate forecasting models are employed. To address this, we introduce AutoMixer, a time-series Foundation Model (FM) approach, grounded on the novel technique of channel-compressed pretrain and finetune workflows. AutoMixer leverages an AutoEncoder for channel-compressed pretraining and integrates it with the advanced TSMixer model for multivariate time series forecasting. This fusion greatly enhances the potency of TSMixer for accurate forecasts and also generalizes well across several downstream tasks. Through detailed experiments and dashboard analytics, we show AutoMixer's capability to consistently improve the Biz-KPI's forecasting accuracy (by 11-15\%) which directly translates to actionable business insights.
\end{abstract}

\section{Introduction}
\label{sec:Introduction}

Business and IT Observability (BizITObs) is an emerging field that focuses on seamlessly integrating business and IT data for comprehensive analysis. In general, the effectiveness of business process execution is assessed using quantifiable metrics known as Business Key Performance Indicators (Biz-KPIs) (e.g. page views, search calls, number of successful or failed transactions). These are captured using Process Monitoring (BPM) tools~\cite{aalst:2004} and are gathered periodically as data streams, offering insights into business performance. Simultaneously, Application Performance Monitoring (APM) tools like Instana\footnote{\url{instana.com}} and Dynatrace\footnote{\url{dynatrace.com}} capture anomalous IT events, such as hardware failures and resource over-utilization as timestamped event logs. The combination of these two data streams, which encompass IT events and Biz-KPIs, constitutes what we term as \biz~data. \biz~data can be structured as a multivariate time-series dataset, wherein each time-point encompasses both the presence and counts of various IT events with its corresponding business KPI values. This formulation allows for a comprehensive analysis of the interplay between IT events and business performance over time.


In this ever-changing ecosystem, where anomalous IT events can swiftly affect business KPIs\footnote{Business KPIs aka Biz-KPIs}, there is a compelling need for accurate multivariate forecasting of these Biz-KPIs.
These models must rapidly and precisely forecast the business KPIs by considering the joint effect of historical IT events and past Biz-KPIs. By harnessing the power of these predictive models, organizations can accurately estimate the potential business impact of various IT failures and promptly take corrective actions before they escalate into significant issues. 
Also, users can prioritize the remediation actions that will have the most significant positive effect on revenue generation and operational efficiency. Thus, accurate forecasting of Biz-KPIs is of utmost importance.


One of the key distinguishing features in  BizITObs time series data lies in its strong correlation across channels where each channel signifies an IT event or business KPI. 
The channels can exhibit different correlation patterns such as IT events with business KPIs (e.g., disk failure affecting payment gateway calls), IT-event correlations (e.g., disk failure impacting database), and business KPI correlations (e.g., throughput linked to cycle time and revenue). 
To achieve precise forecasting in these scenarios, models that effectively target cross-channel correlations are of utmost importance. In this paper, we focus on the problem of accurate cross-channel aware multivariate forecasting of business KPIs given historical data on IT events and Biz-KPIs. 

\begin{figure*}[t]
     \centering
     \begin{subfigure}[b]{0.55\textwidth}
         \centering
         \includegraphics[width=\columnwidth]{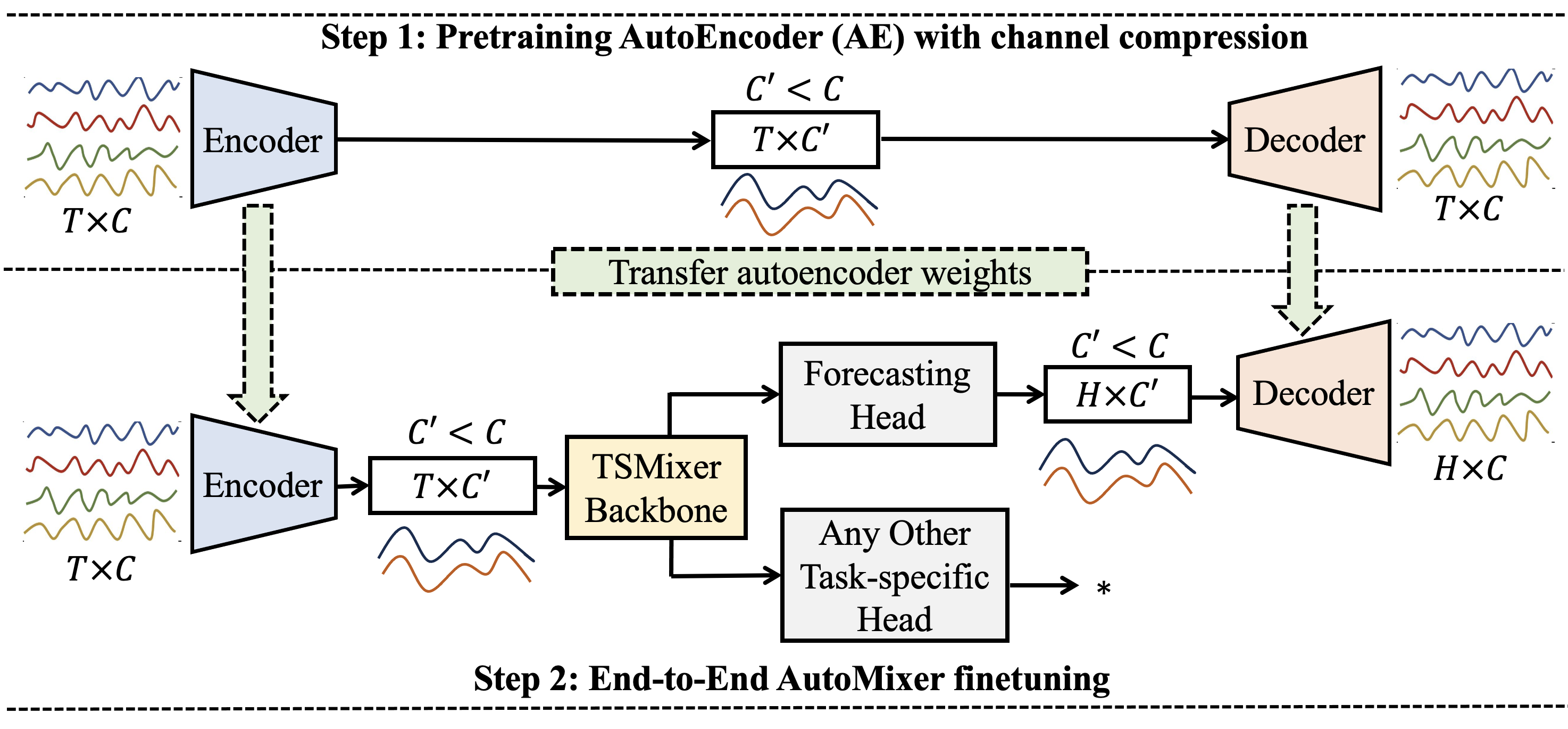}
         \caption{The AutoMixer framework. $T$: input context/history length, $H$: forecast horizon, $C$: number of input channels, $C'$: number of compressed channels.}
         \label{fig:high-level}
     \end{subfigure}
     \hfill
     \begin{subfigure}[b]{0.43\textwidth}
         \centering
         \includegraphics[width=\columnwidth]{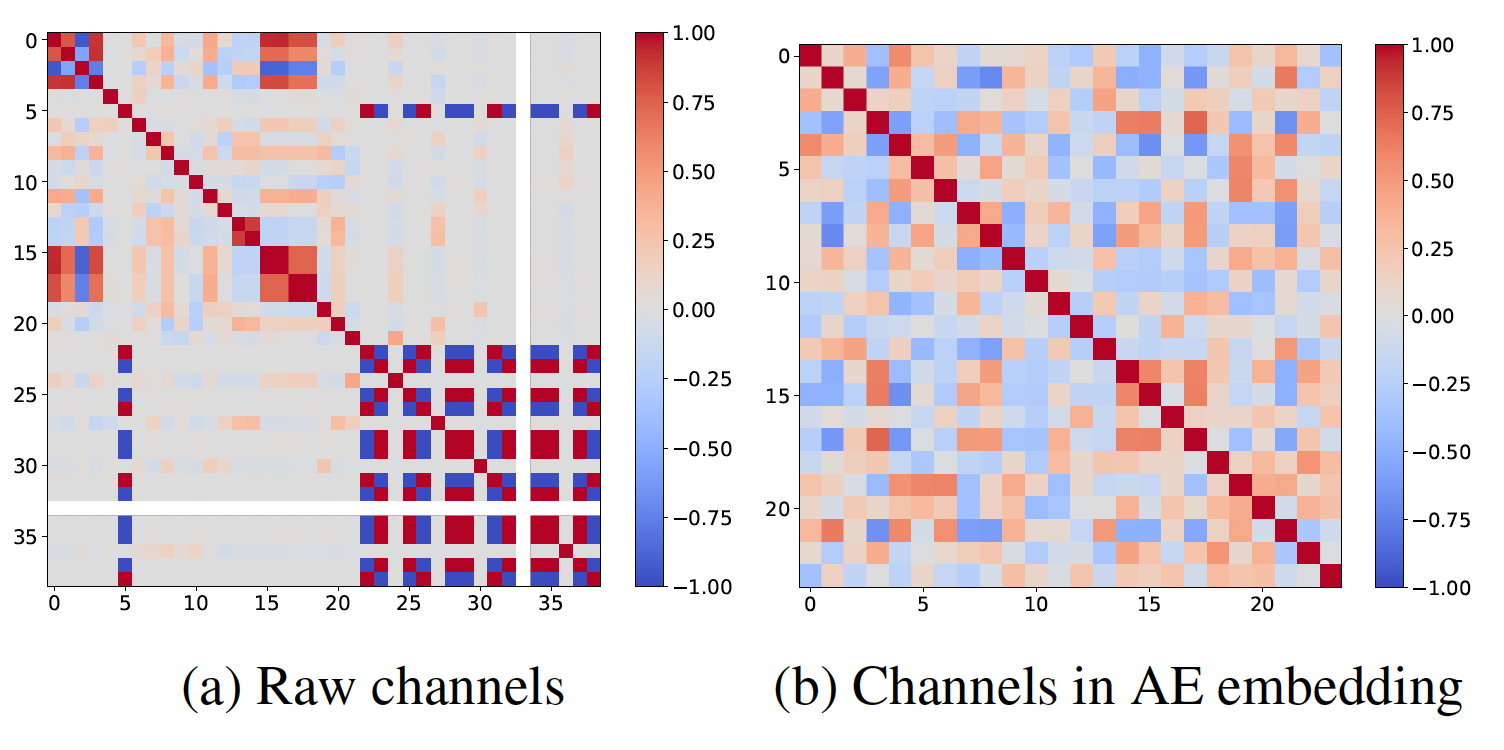}
         \caption{Inter-channel correlation for Application data}
         \label{fig:corr}
     \end{subfigure}
    \caption{
    AutoMixer overview
    }
    \label{fig:amixer}
\end{figure*}
\textbf{Related work:} This is a classic forecasting problem with a well-established line of research dedicated to solving it.
Recent advancements in Transformer~\cite{vaswani2017attention} based forecasting models have surpassed the traditional statistical and ML models~\cite{deepar}~\cite{nbeats} with very high state-of-the-art accuracy. Specifically, Informer~\cite{informer}, Autoformer~\cite{autoformer}, and Fedformer~\cite{fedformer} introduced channel mixing, wherein channels from the same timepoint are combined into an embedding and processed using self-attention networks for cross-channel modeling. These methods propose various improvements to the self-attention modules and were considered state-of-the-art in multivariate forecasting until DLinear~\cite{dlinear} and PatchTST~\cite{patchtst} presented the power of the channel-independence approach. In channel-independence, authors treat multivariate modeling as a global univariate approach with shared model weights across channels, demonstrating superior forecasting accuracy compared to existing channel-mixing techniques. In contrast, CrossFormer~\cite{crossformer} proposed an alternative native cross-channel modeling approach. Further advancements were made with TSMixer~\cite{vijay2023tsmixer}, which introduced the cross-channel reconciliation head for explicit cross-channel modeling, leading to even better forecasting accuracy. To the best of our knowledge, TSMixer is the latest state-of-the-art for multivariate forecasting with improved cross-channel correlation support.

However, when these models were used on BizITObs data, there was evident potential to enhance cross-channel correlations for better accuracy.
Toward this, we propose \textbf{\textit{AutoMixer}}, a novel time-series foundation modeling approach for BizITObs data that harnesses the combined capabilities of AutoEncoder and TSMixer for improved multivariate time-series forecasting. AutoMixer follows a foundation model workflow, wherein the first step involves pretraining an AutoEncoder (AE) independent of the downstream tasks, to compactly project the input data into a compressed channel space and enable full reconstruction. 
Then, for every downstream task (e.g. Biz-KPI forecasting), we seamlessly interleave the pretrained AutoEncoder components at different points in the TSMixer workflow as depicted in Figure~\ref{fig:high-level} and do end-to-end finetuning. 
The intuition behind AutoMixer is elegantly simple. 
Rather than feeding raw channels directly to TSMixer, we harness the power of \textit{compressed channels} (via AutoEncoders~\cite{murphy2022probabilistic}) where unwanted noisy channel interactions are reduced. Hence, the channel correlation task is simplified and the potency of TSMixer is significantly amplified in the compressed channel space, resulting in even more accurate forecasts. 

This comprehensive two-step training process (i.e. pretrain and finetune) allows the framework to unlock the full potential of both AutoEncoder and TSMixer. It also enables the common pretrained models to be easily repurposed for various downstream tasks (such as forecasting, classification, regression, \etc) just by changing the final heads, and harnessing the power of foundation model workflows. 
Note that, in time-series, varying data dynamics mandate separate pretrained models for each dataset, capable of generalizing across multiple downstream tasks within the same dataset~\cite{patchtst, vijay2023tsmixer}. This contrasts with NLP and image domains, where pretrained models are constructed across several datasets.

\subsubsection{Summary of Contributions}
\begin{itemize}
   

\item We introduce \textbf{AutoMixer, a time-series foundation model approach} particularly targeted for BizITObs data for improved Biz-KPI multivariate forecasting via the proposed novel technique of channel-compressed pretrain and finetune workflows. 

\item AutoEncoders are traditionally used for compressing hidden features. In contrast, we employ AutoEncoders to compress the channels instead of hidden features. This simplifies the channel correlation task which in turn amplifies the potency of TSMixer as it works in the compressed less noisy channel space, resulting in even more accurate and insightful forecasts. 

\item We publicly release 4 new BizITObs datasets and provide benchmarks for various state-of-the-art (SOTA) forecasting models. Our comprehensive experiments emphasize AutoMixer's effectiveness, surpassing SOTAs in the Biz-KPI forecasting task with a substantial 11-15\% MSE reduction and 1-2X enhancements in PCC.


\item We also demonstrate the effectiveness and ease of porting these channel-compressed pretrained models across several \biz~downstream tasks, harnessing the power of foundation model workflows.


\item Through an actionable business insight dashboard and real-time examples, we demonstrate how these forecasting improvements practically empower business and IT analytics for effective decision-making. 


\end{itemize}

\section{AutoMixer: A Foundation Model Approach for BizITObs}

\subsection{Problem formulation}
We denote a multivariate time series as $\x \in \mathbb{R}^{T\times C}$, where $T$ is the length of the time series and $C$ is the number of channels (or variates). Multivariate time series data in the BizITObs domain blend channels from both business KPIs and IT events together. The multivariate forecasting problem can be formulated as a sequence-to-sequence prediction problem:
given a history sequence $\x \in \mathbb{R}^{T\times C}$ we have to predict the subsequent future sequence $\y \in \mathbb{R}^{H\times C}$ with a trainable model $f(\cdot)$.
The model's forecast is denoted as $\yhat = f\left( \x \right) \in \mathbb{R}^{H\times C}$.
$H$ is the forecast horizon.

\subsection{TSMixer background}
\label{tsmixer}
Given AutoMixer is built on top of TSMixer, we provide an overview of TSMixer here. TSMixer employs MLPMixer blocks and segments input time series into fixed windows, followed by gated MLP transformations and permutations for enhanced accuracy. TSMixer adopts a channel-independent strategy, modeling all channels globally in a univariate manner, with shared model weights across channels to implicitly account for channel correlations. To explicitly capture channel correlations in a pure multivariate fashion, the TSMixer-CC variant is further introduced, which blends a channel-independent backbone with explicit channel-reconciliation heads. 
This approach achieves the latest state-of-the-art accuracy. Next, we explain how its potency can be further improved using AutoMixer’s channel-compressed pretrain and finetune workflows.

\subsection{Channel compression as a pretraining task}
Foundation models and self-supervised representation learning have been found to be effective in natural language and vision domains.
Pretraining through a self-supervised objective can learn a generic representation of the data, and the pretrained model weights can be subsequently finetuned for various downstream tasks like forecasting or classification.


BizITObs data in general exhibit a strong presence of intricate and potentially lagged correlation patterns within and across business KPIs and IT events. Some of these correlations hold significant importance for effective modeling, while others merely contribute noise and hinder the modeling process. 
APM tools like Instana can potentially capture hundreds of IT events, some of which may be rare. 
It is difficult to determine a priori which IT events are responsible for influencing the business KPIs. 
The IT events which do not affect the business KPIs only add noise to the data.

To tackle this, we propose to project the raw channels to a compressed channel space where unimportant and irrelevant channel correlations are pruned away and only the important correlations are compactly retained.
We pose the channel compression as a pretraining task, and achieve it with an AutoEncoder (AE) model.
Specifically, the AutoEncoder $g(\cdot)$ takes a time series segment $\x \in \mathbb{R}^{T\times C}$ and the encoder part compresses \textit{only} the channel dimension \ie $\z = g\left( \x \right) \in \mathbb{R}^{T\times C'}$, where $C' < C$ is the number of channels in the compressed space. 
The decoder of the AE tries to reconstruct the original $\x$ from the compressed representation $\z$. The decoder's output $\hat{\x} = h(\z) \in \mathbb{R}^{T\times C}$ is compared with the original input $\x$, and the mean squared error (MSE) objective $||\x - \hat{\x}||_2^2$ is employed to train the AE.
Intuitively, if $\x$ can be reconstructed with low MSE, then, the original input data possibly has latent correlations between the channels which was essentially leveraged by the AE for a successful channel compression task. Figure~\ref{fig:corr} shows the inter-channel correlations for raw channels and channels in the embedding space on Application dataset (details in Section~\ref{sec:Data}). We can see that the raw channels have strong absolute correlations between many channels, which is compressed and evenly distributed by the AE. As we observe, these compressed channels demonstrate diminished inter-channel correlation, signifying that important and meaningful correlations are already amalgamated within separate channels in the compressed space. This streamlined structure simplifies the channel correlation task and the potency of TSMixer is significantly amplified in the compressed channel space, resulting in even more accurate and insightful forecasts.



\subsection{End-to-End forecasting as a finetuning task}
The pretrained AE is used in the final stage of training the AutoMixer end-to-end for a downstream Biz-KPI forecasting task.
The input $\x$ is first compressed into $\z$ through the encoder part of the AE.
The backbone of the TSMixer takes the compressed representation $\z \in \mathbb{R}^{T \times C'}$ and converts it into $\zprime \in \mathbb{R}^{C'\times n \times l}$ by a series of transformations. Here, $n$ denotes the number of patches (TSMixer performs the patching internally), and $l$ is the length of every patch. Hence, $T = n\times l$.
The output of the TSMixer backbone can be fed to \textit{task-specific heads}.
For forecasting, the head converts $\zprime$ into $\yprime \in \mathbb{R}^{H\times C'}$ which is fed into the pretrained decoder of the AE to produce the final output, $\yhat \in \mathbb{R}^{H\times C}$.
The entire model is trained (hence, the AE weights are finetuned and the TSMixer weights are trained from scratch) end-to-end with a mean squared objective, $||\y - \yhat||_2^2$.
For any other downstream task like classification or regression, the decoder part of the AE might be removed, and the output of the TSMixer backbone $\zprime$ can be directly fed to the task-specific head to generate the prediction (see Figure~\ref{fig:high-level}). 

\subsection{Discussion on  Design Choices}
In this section, we delve into the intricacies of the AutoMixer architectures, shedding light on the multitude of design choices that have been considered and thoughtfully incorporated. Each component of the AutoMixer has been crafted with a specific purpose in mind, guided by a coherent set of reasons that harmoniously converge to yield an efficient and effective prediction system for ~\biz~ data.

\subsubsection{Choice of AutoEncoders} Autoencoders can be constructed using diverse architectures like CNNs and RNNs. Our forecasting pipeline requires sequence-invariant Autoencoders, allowing flexible handling of different input and output sequence lengths. Since CNNs lack sequence-invariance, we opted for RNN-based Autoencoders. We will explore the contrast between GRU and LSTM in subsequent experiments.


\subsubsection{Choice of Backbone models}  Our approach is adaptable to various deep learning prediction backbones. While options like Informer, Fedformer, Patchtst, and TSMixer were available, TSMixer stood out with its high state-of-the-art accuracy complimented with 2-3X speed and memory improvement. Thus, we chose TSMixer as our preferred backbone among the alternatives.

\subsubsection{Choice of training workflows} 
In training workflows, we face two distinct choices. 
The first approach trains a single AutoMixer end-to-end by initializing the AE weights at random for channel compression. 
Alternatively, we can also follow a two-step process (as done in foundation models~\cite{foundation}): commencing with AE weight learning in a pretraining phase and then transferring this knowledge to enhance the AutoMixer through finetuning. 
Based on empirical analysis (Section~\ref{sec:Experiments}), we choose foundation modeling over direct end-to-end training due to its significant performance improvements.

\begin{table}[]
\small
    \scalebox{0.94}{
    \centering
    \begin{tabular}{|c |c| c| c| c|}
        \hline
        Dataset & \#Obs & \#Channels & \#Biz-KPIs & \#IT Events \\\hline
        Service & 8835 & 107 & 72 & 35 \\
        Application & 8834 & 39 & 4 & 35 \\
        L2C & 31968 & 9 & 7 & 2 \\
        SO & 20475 & 45 & 13 & 32 \\
        \hline
    \end{tabular}}
     \caption{Dataset statistics}
    \label{tab:data_stats}
\end{table}

\begin{table*}[]
\small
\centering
\begin{tabular}{|c|c|cc|cc|cc|cc|}
\toprule
& Datasets                         & \multicolumn{2}{c|}{Application}                       & \multicolumn{2}{c|}{SO}                         & \multicolumn{2}{c|}{Service}                          & \multicolumn{2}{c|}{L2C}                               \\ \hline
Category & Models                           & \multicolumn{1}{c|}{MSE $\scriptstyle \downarrow$}           & PCC  $\scriptstyle \uparrow$             & \multicolumn{1}{c|}{MSE $\scriptstyle \downarrow$}           & PCC  $\scriptstyle \uparrow$             & \multicolumn{1}{c|}{MSE $\scriptstyle \downarrow$}          & PCC  $\scriptstyle \uparrow$             & \multicolumn{1}{c|}{MSE $\scriptstyle \downarrow$}           & PCC  $\scriptstyle \uparrow$             \\ \hline
\multirow{4}{1.6cm}{\textbf{AutoMixer variants (ours)}} & AutoMixer GRU                    & \multicolumn{1}{c|}{0.1535}          & 0.9086          & \multicolumn{1}{c|}{\textbf{0.5627}} & {\underline{0.2198}}    & \multicolumn{1}{c|}{\textbf{0.148}} & \textbf{0.8684} & \multicolumn{1}{c|}{0.0774}          & 0.9719          \\
& AutoMixer GRU CC                 & \multicolumn{1}{c|}{\underline{0.0937}}          &  \underline{0.9354}          & \multicolumn{1}{c|}{0.5697}          & 0.213         & \multicolumn{1}{c|}{\underline{0.1494}}   & {\underline{0.862}}     & \multicolumn{1}{c|}{0.0772}          & 0.9719          \\
& AutoMixer LSTM                   & \multicolumn{1}{c|}{\textbf{0.0788}}    & {\textbf{0.9407}}    & \multicolumn{1}{c|}{0.5694}          & 0.2131          & \multicolumn{1}{c|}{0.1593}         & 0.8569         & \multicolumn{1}{c|}{0.0771}          & 0.9719          \\
& AutoMixer LSTM CC                & \multicolumn{1}{c|}{0.1012} & {0.9324} & \multicolumn{1}{c|}{{0.5657}}          & {0.2194}          & \multicolumn{1}{c|}{0.1602}         & 0.8576          & \multicolumn{1}{c|}{0.0772}          & 0.9719          \\ \midrule
\multirow{5}{1.6cm}{SOTAs} & TSMixer                          & \multicolumn{1}{c|}{0.2194}          & 0.8352          & \multicolumn{1}{c|}{0.6042}          & 0.0096          & \multicolumn{1}{c|}{0.2094}         & 0.8576          & \multicolumn{1}{c|}{0.0723}          & {\underline{0.9734}}    \\
& TSMixer CC                       & \multicolumn{1}{c|}{0.1313}          & 0.9148          & \multicolumn{1}{c|}{\underline{0.5636}}    & 0.1641          & \multicolumn{1}{c|}{0.1654}         & 0.8542          & \multicolumn{1}{c|}{0.0716}          & \textbf{0.9737} \\ 
& PatchTST                          & \multicolumn{1}{c|}{0.3484}                & 0.5735                & \multicolumn{1}{c|}{0.8572}                & 0.0647               & \multicolumn{1}{c|}{0.2392}               & 0.7873               & \multicolumn{1}{c|}{0.1086}                &   0.9574            \\

& FEDFormer                          & \multicolumn{1}{c|}{0.4643}                & 0.4855               & \multicolumn{1}{c|}{0.8971}                & 0.0570              & \multicolumn{1}{c|}{0.2845}               & 0.7524               & \multicolumn{1}{c|}{0.1342}                &   0.9475          \\

& DLinear                          & \multicolumn{1}{c|}{0.3256}                & 0.5983            & \multicolumn{1}{c|}{0.9089}                & 0.0050              & \multicolumn{1}{c|}{0.2536}               & 0.7875             & \multicolumn{1}{c|}{0.1046}                &   0.9578          \\

\midrule
\multirow{6}{1.6cm}{Baselines}

& VAR                              & \multicolumn{2}{c|}{NA}                                & \multicolumn{1}{c|}{0.6505}          & 0.1603          & \multicolumn{2}{c|}{NA}                               & \multicolumn{1}{c|}{1.0100}          & 0.1131          \\
& XGBoost                          & \multicolumn{1}{c|}{2.7192}                & 0.3923                 & \multicolumn{1}{c|}{0.8331}                & 0.1804                & \multicolumn{1}{c|}{0.2955}               & 0.2065                & \multicolumn{1}{c|}{0.1545}                &   0.0574              \\
& AGCRN                            & \multicolumn{1}{c|}{0.2653}          & 0.6458          & \multicolumn{1}{c|}{0.5839}          & \textbf{0.3159} & \multicolumn{1}{c|}{0.1515}         & 0.2491          & \multicolumn{1}{c|}{\textbf{0.0671}}    & 0.1851          \\
& DecoderMLP                       & \multicolumn{1}{c|}{0.5591}          & -0.4585         & \multicolumn{1}{c|}{0.6943}          & 0.0015          & \multicolumn{1}{c|}{0.5892}         & -0.0070         & \multicolumn{1}{c|}{0.1233}          & 0.0585          \\
& Recurrent GRU                    & \multicolumn{1}{c|}{0.4873}          & 0.0763          & \multicolumn{1}{c|}{0.7118}          & 0.1203          & \multicolumn{1}{c|}{0.3326}         & 0.1128          & \multicolumn{1}{c|}{0.092}           & 0.0645          \\
& TFT                              & \multicolumn{1}{c|}{0.3843}          & 0.1698          & \multicolumn{1}{c|}{0.7283}          & 0.0901          & \multicolumn{1}{c|}{0.3435}         & 0.1295          & \multicolumn{1}{c|}{\underline{0.0694}}          & 0.2588          \\ \hline

& \makecell{\textbf{Avg Imp. over best SOTA}} & \multicolumn{8}{c|}{\textbf{MSE: 11\%, PCC: 1X}} \\ \hline
& \makecell{\textbf{Avg Imp. over best Baseline}}   & \multicolumn{8}{c|}{\textbf{MSE: 15\%, PCC: 2X}} \\ \bottomrule


\end{tabular}
\caption{Comparing AutoMixer with benchmarks for Biz-KPIs forecasting task. Comparison is across AutoMixer best variant with the benchmark's best variant in SOTA and Baselines categories. Average Improvement (Imp.) across all datasets \textit{w.r.t} MSE is given in \% and improvement \textit{w.r.t} PCC is in X's (number of times).
    $\scriptstyle \downarrow$ indicates lower is better. $\scriptstyle \uparrow$ indicates higher is better.
    The best and second-best results are in \textbf{bold} and \underline{underlined} respectively. VAR models failed in few datasets due to constant channel values.}

\label{tab: Master_table}
\end{table*}

\section{Experiments and Results}\label{sec:Experiments}
We comprehensively evaluate AutoMixer's accuracy against established benchmarks and perform ablation studies to emphasize the importance of different model configurations.

\subsection{Data}\label{sec:Data}
For experimental purposes, we used four datasets that are synthesized based on various realistic business process workflows. \footnote{We will release the data soon}


\subsubsection{Service} This dataset relates to the ``Stan's Robot Shop"~\cite{robotshop} app on the cloud, overseen by Instana. It models a user's e-commerce journey, from site access to shipping, using a load generator. Intermittent fault injection induces various IT events. It offers business KPIs for services (e.g. payment, catalog) and Instana-tracked IT events. It is sampled every 10 seconds due to high traffic and events.


\subsubsection{Application} This dataset is similar to the above dataset but contains KPIs at the level of the whole application.

\subsubsection{L2C} This dataset is related to the Lead-to-Cash (L2C) process ranging from customers' intent or interest to buy a product, to a company’s realization of revenue based on product sales~\cite{l2c}. It covers steps like order updates, approvals, and acceptance. It is sampled every 5 minutes due to moderate IT event frequency.


\subsubsection{SO} This dataset pertains to the Sales Order (SO) process, encompassing order creation, outbound deliveries, goods issues, and invoicing~\cite{so}. Generated in a controlled setting based on the existing sales order process with additional order validation, and order correction, the data is sampled every 1 hour due to infrequent IT events. It also includes environmental metrics from Environmental Intelligence Suite~\cite{eis}, designed to potentially impact outbound shipping in adverse weather. These metrics are also treated as auxiliary IT events in our experiments.


Table~\ref{tab:data_stats} provides the total number of observations, time series variables, KPIs, and IT events for each dataset.

\subsection{Experimental setting}
\subsubsection{Model Variants}
We present four AutoMixer variants: (i) AutoMixer LSTM, (ii) AutoMixer LSTM CC, (iii) AutoMixer GRU, and (iv) AutoMixer GRU CC. These options arise from different underlying model (TSMixer, TSMixer-CC) and AutoEncoder (LSTM, GRU) selections.



\subsubsection{Forecasting Benchmarks} We compare the AutoMixer variants with various established multivariate time series forecasting benchmarks, primarily grouped as follows: (i) \textbf{Mixer Models:} TSMixer and TSMixer-CC (Latest SOTAs, refer to Section~\ref{tsmixer}), (ii) \textbf{Other Deep Learning Models:} PatchTST (Nai et al. 2022), DLinear (Zeng et al. 2023), FEDFormer ( Zhou et al. 2022),   Recurrent-GRU~\cite{reccurent_gru}, DecoderMLP and TFT~\cite{tft} from the PyTorch forecasting library~\cite{pytorch}, AGCRN~\cite{bai2020adaptive} (iii) \textbf{Other ML and Statistical Models:} XGBoost Regressor~\cite{xgboost} and VAR ~\cite{var}. 


\subsubsection{Model Configuration} 
First, we trained the autoencoder with a context/history length, $sl =T= 24$, and a forecast length, $fl =H= 24$.
Subsequently, we utilized these weights in the AutoMixer for the end-to-end training for forecasting.
Throughout all the datasets and experiments, the following model configurations from TSMixer paper is used: patch length ($pl$) $= 8$, batch size ($b$) $= 8$, feature scalar ($fs$) $= 2$, hidden feature size ($hf$) = $fs \times pl$ $= 16$, and expansion feature size ($ef$) = $fs \times hf$ $= 32$. For SO and Application data, we used the number of Mixer layers ($nl$)$= 8$ and dropout ($do$) $= 0.4$, and for the Service and L2C data, $nl = 3$ and $do= 0.3$. The training, validation, and test sets are chronologically split in a ratio of $0.6:0.2:0.2$ using temporal cross-validation~\cite{hpro}. 
For all training, we incorporated early stopping based on the validation performance.
For each dataset, we choose the optimal channel compression ratio ($cr$) and model variant by assessing their effects on the validation set.

\subsubsection{Metrics}
We use the mean squared error (MSE) to measure the model performance and also employ the Pearson Correlation Coefficient (PCC) to evaluate the correlation between predicted and actual value. PCC is particularly crucial for BizITObs data, as we expect a good model to not only capture mean predictions but also capture accurate trends and patterns.
We independently normalize channels by subtracting means and dividing by standard deviations. Both loss and reported metrics are then computed on this normalized data, eliminating bias from the original data scale.

\begin{table}[]
\small
\centering
\begin{tabular}{|c|cc|}
\hline
Data        & \multicolumn{1}{c|}{PT}                              & NPT                            \\ \hline
Application & \multicolumn{1}{c|}{0.0937}                          & 0.0956                         \\
SO          & \multicolumn{1}{c|}{0.5697}                          & 0.5551                         \\
Service     & \multicolumn{1}{c|}{0.1494}                          & 0.1618                         \\
L2C         & \multicolumn{1}{c|}{0.0772}                          & 0.079                          \\ \hline
Avg \% Imp  & \multicolumn{2}{c|}{2.3\%}                                                            \\ \hline
\end{tabular}
 \caption{MSE Improvements with pretraining (PT) vs  without pretraining (NPT).}
 \label{tab: pretraining}
\end{table}

\begin{table}[t]
\small
\centering
\begin{tabular}{|c|ll|ll|}
\hline
$cr$   & \multicolumn{1}{c|}{Application}  & \multicolumn{1}{c|}{SO}        & \multicolumn{1}{c|}{Service}   & \multicolumn{1}{c|}{L2C}                                \\ \hline
20\% & \multicolumn{1}{l|}{0.0935}          & 0.5697          & \multicolumn{1}{l|}{0.1677}          & \multicolumn{1}{c|}{-} \\
40\% & \multicolumn{1}{l|}{0.1074}          & 0.5715 & \multicolumn{1}{l|}{\textbf{0.1463}} & 0.0819                 \\
60\% & \multicolumn{1}{l|}{0.0937}          & 0.5708         & \multicolumn{1}{l|}{0.1494}          & \textbf{0.0772}        \\
80\% & \multicolumn{1}{l|}{\textbf{0.0731}} & \textbf{0.5598 }         & \multicolumn{1}{l|}{0.1485}          & \multicolumn{1}{c|}{-} \\ \hline
\end{tabular}
\caption{Comparing across channel compression ratios ($cr$). ``-" indicates compression infeasibility due to fewer channels}
\label{tab:compression}
\end{table}

\begin{table}[t]
\small
 \scalebox{0.97}{
\begin{tabular}{|c|c|c|c|c|}
\hline
Datasets                     & Model     & \begin{tabular}[c]{@{}c@{}}Biz-KPI\\ Forecast\\ (MSE $\scriptstyle \downarrow$)\end{tabular} & \begin{tabular}[c]{@{}c@{}}IT Event\\ Forecast\\ (MSE $\scriptstyle \downarrow$)\end{tabular} & \begin{tabular}[c]{@{}c@{}}IT Event\\  Clf\\ (Acc $\scriptstyle \uparrow$)\end{tabular} \\ \hline
\multirow{3}{*}{Application} & AutoMixer & \textbf{0.0937}                                                   & \textbf{0.3035}                                                & \textbf{77.82\%}                                                      \\
                             & TSMixer CC   & 0.1313                                                            & 0.32                                                           & 33.92\%                                                               \\
                             \hline
\multirow{3}{*}{Service}     & AutoMixer & \textbf{0.1494}                                                   & \textbf{0.3255}                                                & \textbf{83.21\%}                                                      \\
                             & TSMixer CC   & 0.1654                                                            & 0.3294                                                         & 44.75\%                                                               \\
                             \hline
    \multicolumn{2}{|c|}{\textbf{Avg improvement}}              & 19.2\%          & 3.2\%          & 2X \\
\hline
\end{tabular}
}
\caption{Comparing downstream tasks,  KPI forecast, IT event forecast and IT event classification (Clf). Avg improvements are \textit{w.r.t} AutoMixer GRU CC over TSMixer CC  on MSE and Accuracy (Acc).}
\label{tab:downstream}

\end{table}

\subsection{Biz-KPI forecasting}
\label{bforecast}
In this section, we delve into various empirical studies highlighting the potential of AutoMixer for Biz-KPI forecasting. All results in this section report the MSE or PCC averaged across all the normalized Biz-KPI channels at a dataset level.

\subsubsection{Accuracy Improvements}

We compare AutoMixer to existing benchmarks for multivariate forecasting of Biz-KPIs ($fl$ = 24), using past IT events and Biz-KPIs ($sl$ = 24). The results presented in Table~\ref{tab: Master_table} reveal AutoMixer's superior performance over Baselines, displaying a 15\% average reduction in MSE and a twofold increase in PCC. When compared to TSMixer and other SOTAs, AutoMixer demonstrates an average MSE improvement of 11\%, though there are no significant PCC improvements. This can be attributed to TSMixer's strong trend-capturing abilities, which limit the potential for notable PCC improvements, despite notable enhancements in MSE.  
These insights highlight the potency of AutoMixer to capture both mean and trends more accurately in most of the considered datasets, a challenge in existing benchmarks. Among the considered AutoMixer variants, each variant performed well in different scenarios, and hence in practice, we would need to select the best variant using the validation metric for customer deployments.

\subsubsection{Ablation Study}

In this section, we study how a couple of crucial design settings, specifically pretraining and channel compression ratios, influence the accuracy of AutoMixer, taking the AutoMixer GRU CC variant as an example.


\begin{figure*}[t]
    \centering
    \includegraphics[width=2\columnwidth]{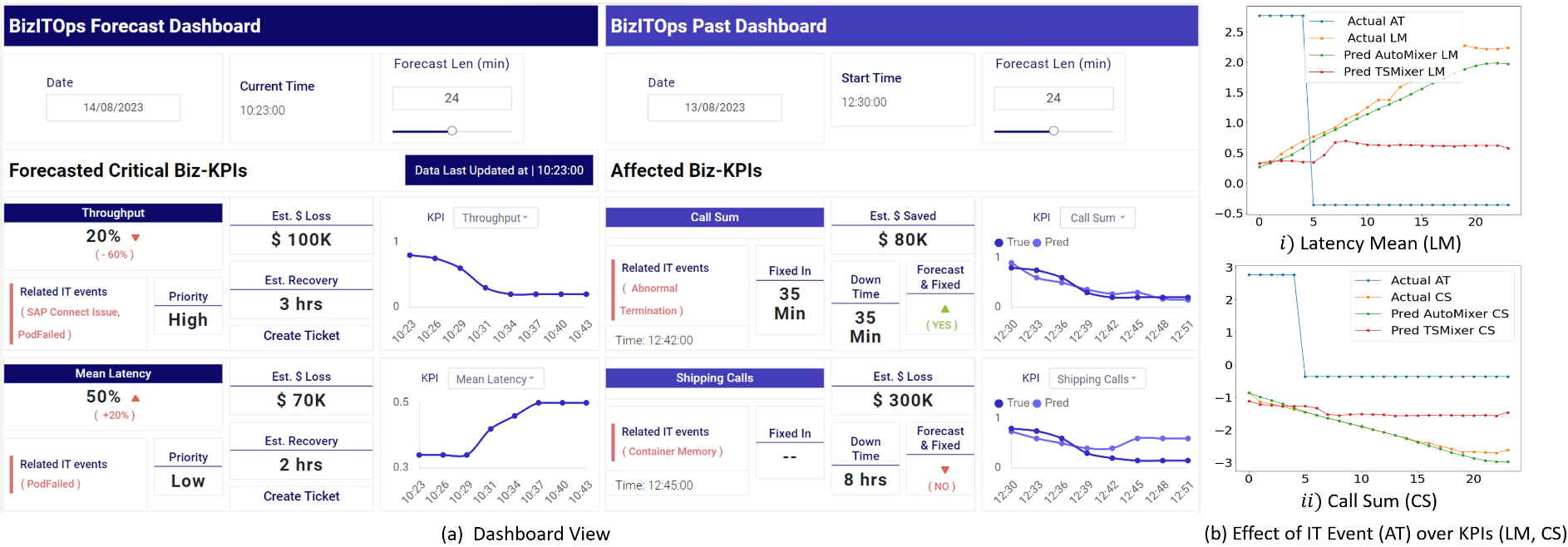}
    \caption{Actionable Business Insight Dashboard and real-time prediction examples
}
    \label{fig:bixview}
\end{figure*}
\textbf{Effect of pretraining:} 
Here, we analyze the impact of pretraining on AutoMixer for Biz-KPI forecasting. Table ~\ref{tab: pretraining} highlights that utilizing pretraining (PT) of AE reduces the MSE by 2.3\% as compared to direct AutoMixer training  with randomly initialized AE weights (NPT). 
This reiterates our initial hypothesis that employing the foundation modeling workflow can provide better generalization and improved forecasts. Moreover, these pretrained models can easily cater to multiple downstream tasks as explained in Section~\ref{downsection}

\textbf{Effect of Channel Compression:} Channel compression ratio is an important hyperparameter for improving AutoMixer's accuracy, and it varies across datasets as shown in the Table~\ref{tab:compression}. Hence based on the validation metric, we choose the right compression ratio for every dataset and model variant. This validation-based evaluation is important to get the right trade-off between channel-compression and information loss for a given model and data. However, in general, the best compression ratios mostly lie in the range between 40-80\% leading to improved model performance. This compression helps us eliminate channels that contribute noise to the model.

\subsection{Applicability to multiple downstream tasks}
\label{downsection}
In this section, we show the applicability of our pretrained model across 3 downstream tasks \viz (i) Biz-KPI forecasting: Temporal forecast of Biz-KPIs for the next 24 points (same as in Section~\ref{bforecast}), (ii) IT event forecasting: Temporal forecast of IT events for the next 24 points (iii) IT event classification: Multi-label classification of various IT events to occur in the next 24 points. As shown in Table ~\ref{tab:downstream},  the channel compressed encoder from the pretrained model when used in conjunction with TSMixer consistently boosts the performance of TSMixer across all the considered downstream tasks. 
Thus, the one-time pretraining step facilitates improved generalization across multiple downstream tasks, highlighting the potential of time-series foundation models.

\section{Actionable Business Insight Dashboard}
In this section, we present the prototype dashboard view that conceptualizes how AutoMixer's forecasting benefits enhance actionable business insights and decision-making. Figure~\ref{fig:bixview}a showcases the \biz~Forecast and Past Dashboards. In the Forecast Dashboard (left view), we display Top-k biz-KPIs with crucial forecasts requiring preemptive attention to avert downtime or revenue loss. The system also estimates potential downtime and revenue loss based on similar historical ticket data, providing these estimates in the UI for informed decisions. It also assigns a priority score for each insight for effective ranking during resource constraints. In the Past Dashboard (right view), we exhibit the top-k affected Biz-KPIs within a defined timeframe and indicate whether AutoMixer has forecasted it correctly or missed it. This dashboard also highlights the impact of a forecast hit or miss by visually representing the downtime prevented/incurred and revenue saved/lost estimated from similar historical past tickets. Figure~\ref{fig:bixview}b (i, ii) illustrates an actual forecasting instance involving 2 Biz-KPIs [Calls Sum (CS), Latency Mean (LM)] in response to an IT event [Abnormal Termination (AT)]. Notably, AutoMixer demonstrates superior accuracy compared to TSMixer in both scenarios, offering valuable preemptive business insights and informed data-driven decisions when presented via actionable business insight dashboards.


\section{Conclusion}
In summary, we introduce AutoMixer, a novel time-series foundational model designed specifically for \biz~data. Through comprehensive empirical analysis, we demonstrate the efficacy of our proposed channel compression based pretrain and finetune workflows in capturing the combined impact of historical IT events and past Biz-KPIs. This technique leads to improved Biz-KPI forecasting accuracy (by 11-15\%), and also generalizes well to multiple downstream tasks. We then showcase how these forecasting insights translate into valuable business insights through actionable dashboards, facilitating swift and informed decision-making. Our next steps involve deploying AutoMixer in select customer ecosystems, conducting real-time user studies, and analyzing deployment outcomes.

\bibliography{aaai24}

\end{document}